\documentclass{article}

\usepackage{PRIMEarxiv}

\usepackage[utf8]{inputenc} 
\usepackage[T1]{fontenc}    
\usepackage{hyperref}       
\usepackage{url}            
\usepackage{booktabs}       
\usepackage{amsfonts}       
\usepackage{nicefrac}       
\usepackage{microtype}      
\usepackage{lipsum}
\usepackage{fancyhdr}       
\usepackage{graphicx}       
\graphicspath{{media/}}     
\usepackage{enumerate}
\usepackage{mathtools}
\usepackage{float}
\usepackage{pifont}
\usepackage{subcaption}

\usepackage{amsopn, amssymb, amscd, amsmath, amsthm}

\usepackage{tikz-cd}

\usepackage{physics}

\usepackage{bbm}

\newcommand{\argmin}{\mathop{\mathrm{argmin}}}

\DeclareMathOperator{\IG}{IG}

\providecommand{\R}{\mathbb{R}}
\providecommand{\C}{\mathbb{C}}

\providecommand{\GL}{\mathrm{GL}}

\providecommand{\SO}{\mathrm{SO}}

\providecommand{\Aff}{\mathrm{Aff}}

\providecommand{\Id}{\mathrm{Id}}
\providecommand{\dd}{\mathrm{d}}
\providecommand{\exp}{\mathrm{exp}}

\providecommand{\lie}{\mathfrak}

\providecommand{\gl}{\mathfrak{gl}}

\providecommand{\On}{\mathrm{O}}

\usepackage{algorithm}
\usepackage{algpseudocode}

\providecommand{\Sym}{\mathrm{Sym}}

\theoremstyle{plain}
\newtheorem{theorem}{Theorem}[section]
\newtheorem{proposition}[theorem]{Proposition}
\newtheorem{corollary}[theorem]{Corollary}

\theoremstyle{definition}
\newtheorem{definition}[theorem]{Definition}

\theoremstyle{remark}
\newtheorem{remark}[theorem]{Remark}

\newtheorem{example}[theorem]{Example}

\providecommand{\Id}{\mathrm{Id}}
\providecommand{\dd}{\mathrm{d}}
\providecommand{\exp}{\mathrm{exp}}
\providecommand{\lie}{\mathfrak}
\providecommand{\R}{\mathbb{R}}

\providecommand{\GL}{\mathrm{GL}}
\providecommand{\gl}{\mathfrak{gl}}

\providecommand{\Sym}{\mathrm{Sym}}

\providecommand{\C}{\mathbb{C}}

\providecommand{\SO}{\mathrm{SO}}

\usepackage[normalem]{ulem}

\newcommand{\stkout}[1]{\ifmmode\text{\sout{\textcolor{red}{\ensuremath{#1}}}}\else\sout{\textcolor{red}{#1}}\fi}

\title{Algebraic Adversarial Attacks on Explainability Models}

\author{
  Lachlan Simpson$^{1}$, Federico Costanza$^{2}$, Kyle Millar$^{3}$, Adriel Cheng,$^{1,3}$\\ \textbf{Cheng-Chew Lim}$^{1}$\textbf{, and Hong Gunn Chew}$^{1}$\\
  $^{1}$School of Electrical and Mechanical Engineering, The University of Adelaide, Australia \\
  $^{2}$School of Computer and Mathematical Sciences, The University of Adelaide, Australia\\
  $^{3}$Information Sciences Division, Defence Science \& Technology Group, Australia \\
  \texttt{\small\{lachlan.simpson, esteban.costanza, honggunn.chew, adriel.cheng, cheng.lim\}@adelaide.edu.au}\\
  \texttt{\small\{kyle.millar1, adriel.cheng\}}@defence.gov.au
}

\begin{document}
\maketitle

\begin{abstract}
   Classical adversarial attacks are phrased as a constrained optimisation problem. Despite the efficacy of a constrained optimisation approach to adversarial attacks, one cannot trace how an adversarial point was generated. In this work, we propose an algebraic approach to adversarial attacks and study the conditions under which one can generate adversarial examples for post-hoc explainability models. Phrasing neural networks in the framework of geometric deep learning, algebraic adversarial attacks are constructed through analysis of the symmetry groups of neural networks. Algebraic adversarial examples provide a mathematically tractable approach to adversarial examples. We validate our approach of algebraic adversarial examples on two well-known and one real-world dataset. 
\end{abstract}


\section{Introduction}

Neural networks have become state-of-the-art solutions to a wide array of tasks from computer vision \cite{yolo}, natural language processing \cite{Liu_2023}, cyber-security \cite{9737249}, to fundamental problems in biology \cite{alphafold}. An unsolved problem in machine learning is how neural networks achieve their high performance \cite{Sejnowski_2020}. The limited understanding of how neural networks learn is known as the black-box problem \cite{zednik2019solving}. The black-box problem becomes critical as neural networks have become prevalent within safety critical systems such as self-driving cars, medicine and cyber-security. Post-hoc explainability models are methods for explaining the features that influence the output of a neural network in a human understandable format \cite{ieeeXAIsurvey}. Post-hoc explainability is a step towards addressing the black-box problem \cite{zednik2019solving}. 

Neural networks have been shown to be susceptible to adversarial attacks in applications that include facial recognition \cite{dong2019efficient} and self driving cars \cite{8578273}. Neural network exploits were achieved by observing that a small change in the input resulted in a large change in the output \cite{szegedy2014intriguing}. Adversarial learning on neural networks has highlighted the drastic consequences that a compromised neural network can have \cite{cnn_attack}. Likewise, adversarial learning on explainability models provides orthogonal reasoning used by the neural network to make a classification \cite{xai_attack}. Adversarial explainability is typically phrased as a constrained optimisation problem \cite{{Ghorbani_Abid_Zou_2019,NEURIPS2019_bb836c01,adv_perturb,BANIECKI2024102303}}. Despite the efficacy of a constrained optimisation approach to adversarial attacks, one cannot trace how an adversarial point was generated. The only information available is the point solution to the optimisation problem.

In this work, we take an algebraic approach to adversarial explainability. In \cite{bronstein2021geometricdeeplearninggrids} Bronstein et al. introduce geometric deep learning (GDL). GDL realises that all deep learning models follow a similar architecture that respects the geometry of the underlying input domain. Through studying the symmetry groups of the input domain, one can construct deep learning models purpose built for the input. Lie groups and algebras are fundamental to the study of symmetry and reveals fundamental properties of functions. In this work, Lie theory is applied to study the conditions under which one can exploit algebraic properties of a neural network to produce adversarial explanations. Our work demonstrates that neural networks have fundamental properties leaving them vulnerable to adversarial attacks. Furthermore an algebraic approach to adversarial attacks provides a mathematically tractable process to generating an adversarial attack.

The contributions of this work are threefold:
\begin{enumerate}
    \item We propose algebraic adversarial attacks, a mathematically tractable approach to adversarial examples. The framework of geometric deep learning is used to provide a rigorous theoretical investigation into adversarial attacks.
     
    \item We demonstrate how to use Lie theory to compute algebraic adversarial attacks for base-line attribution methods, additive attribution methods and neural conductance. Our main results are presented in Theorems      \ref{prop:mult_ig_attack}, \ref{prop:SG_attack} and \ref{thrm:alg_attack_lime}.

    \item We demonstrate the efficacy of algebraic adversarial attacks on one real-world and two well-known datasets.
\end{enumerate}

The remainder of this work is structured as follows: Section 2 introduces related work on adversarial explainability and geometric deep learning. In Section 3 we introduce the notion of algebraic adversarial examples. Furthermore, we demonstrate with complete generality that it is always possible to choose examples that satisfy a fixed error threshold. In Section 4 we provide an algebraic description of the symmetries of neural networks via geometric deep learning. We demonstrate how to compute the group of symmetries via the exponential map of the Lie algebra. In Section 5 we theoretically demonstrate how to compute adversarial examples for path-based attribution methods, neural conductance and additive attribution methods. In Section 6 we experimentally validate our approach on one real-world and two well-know datasets. We conclude in Section 7 with a discussion for future work.

\section{Related Work}

\subsection{Adversarial Explainability}

Adversarial examples were first proposed by Ghorbani et al. \cite{Ghorbani_Abid_Zou_2019}. Finding adversarial examples are formulated as a constrained optimisation problem. The input is iteratively perturbed in the gradient direction of the neural network \cite{BANIECKI2024102303} subject to the constraints that 1) the perturbation and input have the same classification and 2) the difference between the input and perturbation are small. Constrained optimisation is the standard approach to adversarial examples \cite{BANIECKI2024102303,9879952,wang2022featureimportanceawaretransferableadversarial,mandal2023defenseadversarialattacksusing}  where the loss function is altered whilst retaining the constraints introduced Ghorbani et al to produce adversarial examples.

Dombrowski et al. \cite{NEURIPS2019_bb836c01}  take a different approach where an adversarial example is optimised to be a specific explanation. Dombrowski et al. further demonstrate that the vulnerability of an explainability model is linked to the large curvature of the output manifold of the neural network. We note that Domrowski's work still meets the constraints of Ghorbani. 

Zhang et al. \cite{Zhang_2022} propose adversarial transferability attacks via neuron attribution. Zhang et al. suppose the architecture of the neural network under attack is unknown except for the features and output. A surrogate model is trained under the same features and adversarial attacks are generated for the surrogate model. Zhang et al. demonstrate that the adversarial attacks on the surrogate model can be transferred to the target model.

Adversarial patches proposed by Brown et al. \cite{brown2018adversarialpatch} add a small patch to the image such that it is misclassified by the model. Nandy et al. \cite{9207312} extend this approach to adversarial examples for explainability models. Nandy et al. add an adversarial patch such that the classification is unchanged but the explanation is drastically different.

The central theme of related work is to minimise a loss function to produce close by points with the same classification, but drastically altered explanations \cite{adv_perturb,BANIECKI2024102303}. Our work is the first to take an algebraic approach to the problem of adversarial explanations. We assume, as with other work, that one has access to the neural network under attack \cite{goodfellow2015explainingharnessingadversarialexamples, kurakin2017adversarialexamplesphysicalworld, madry2019deeplearningmodelsresistant}.

\subsection{Geometric Deep Learning and Explainable AI}
Bronstein et al. \cite{bronstein2021geometricdeeplearninggrids} introduce the geometric deep learning (GDL) framework. GDL realises that most input domains (e.g. images and graphs) have a specific symmetry group acting on the domain  (e.g. translations $\R^{n}$, rotations $\SO_{n}$ and permutations $S_{n}$). Noticing common domains have a specific symmetry group, one can construct neural networks that respect the underlying algebraic properties of the input domain. Atz et al. \cite{Atz2021} note that despite the efficacy of GDL there is little application of explainable AI to GDL architectures. To the best of the authors knowledge this is the first work using the GDL framework to design adversarial explainability attacks. 

The goal of this work is to exploit algebraic properties of a neural network to construct adversarial examples for explainability models. Lie groups of neural networks are studied to explore such properties. Lie groups are fundamental algebraic objects to study symmetry. Whilst we assume familiarity with Lie theory, we provide the necessary basic definitions and results in Appendix \ref{appendix:lie-alg}. Here, and throughout the paper, Lie groups are matrix Lie groups unless otherwise stated.

\section{Classical and Algebraic Adversarial Attacks}
 
\begin{definition}
\label{def:adv_attack_classical}
Given the triple $(\phi,F,x)$, where $\phi$ is an explainability model, $F \in F(\R^{n}, \R^{m})$ is a neural network, $x \in \R^n$ is a point, and an error threshold $\varepsilon >0$. A point $\Tilde{x}$ is an adversarial attack on $(\phi,F,x)$ if it satisfies the following conditions:
\begin{enumerate}
    \item $\|\Tilde{x} - x\|_{\infty} \leq \varepsilon$.
    \item $F(\Tilde{x}) = F(x)$.
    \item $\phi(F,\tilde{x}) \neq \phi(F, x)$.
\end{enumerate}
\end{definition}
Finding a point $\Tilde{x}$ that satisfies the conditions of Definition \ref{def:adv_attack_classical} is solved by a constrained optimisation problem of the following form:
\begin{equation*}
    \argmin_{\Tilde{x}} \mathcal{L}(\Tilde{x},x) \quad \text{ s.t. } \quad \|\Tilde{x} - x\|_{\infty} \leq \varepsilon,
\end{equation*}
where the loss function $\mathcal{L}$ ensures the conditions of Definition \ref{def:adv_attack_classical} are satisfied.

Following the definition of algebraic adversarial attacks for integrated gradients proposed in \cite{alg_attacks_icmlc}, we define algebraic adversarial attacks with complete generality.

		\begin{definition}
            \label{def:alg_attack}
			Let $\tilde{x} \in \R^{n}$ be an adversarial attack on $(\phi, F, x)$. We will say that $\tilde{x}$ is \textit{algebraic} if there exists a group $G$ acting on $\R^{n}$, and an element $g \in G$, such that $\tilde{x} = g \cdot x$.
        \end{definition}
        
        The conditions for a point $\tilde{x} = g \cdot x$ to be an adversarial attack, as per Definition \ref{def:adv_attack_classical}, become the following conditions to be satisfied by $g$:
            \begin{enumerate}
                \item $\|g \cdot x - x\|_{\infty} \leq \varepsilon$.
                \item $F(g \cdot x) = F(x)$.
                \item $ \phi(F, g \cdot x) \neq \phi(F, x)$.
            \end{enumerate}

    \begin{remark}
    All algebraic adversarial attacks are classical adversarial attacks, by definition. Furthermore, $\R^{n}$ could be replaced with any arbitrary vector space.
    \end{remark}

    The remainder of this work will be concerned with computing the group $G$, for a given neural network, that satisfies the conditions of Definition \ref{def:alg_attack}. In Section 5 we demonstrate how to achieve condition 2), for several classes of neural networks. Section 6 will demonstrate how condition 3) holds for additive and path-based attribution methods. The following proposition demonstrates that condition 1) holds with complete generality for an arbitrary Lie group $G$.

  \begin{proposition}\label{prop:threshold}
      Let $G$ be a matrix Lie group acting on $\R^{n}$, let $\lie{g}$ be its Lie algebra and let $x \in \R^{n}$. Then for an arbitrary $\varepsilon \geq 0$ there exists $g 
      \in G$, $g \neq \Id$, such that the following inequality holds:
      \begin{equation}\label{eqn:bound}
          \|g \cdot x - x\|_{\infty} \leq \varepsilon.
      \end{equation}
      Moreover, $g = \exp(t\mathbf{A})$ satisfies equation (\ref{eqn:bound}) for all $t$ such that
      \begin{equation*}
          |t| \leq \frac{1}{\|\mathbf{A} \|_{\infty}}\log \left (\frac{\varepsilon}{\|x\|_{\infty}} + 1 \right),
      \end{equation*}
      with $\mathbf{A} \in \lie{g}$.
  \end{proposition}
  \begin{proof}
      Let $\mathbf{A} \in \mathfrak{g}$ and $g_{t} = \exp(t\mathbf{A}) \in G$ for $t \in \R$. From the definition of the exponential of matrices, we observe that
      \[
        g_{t} - \Id = \sum\limits_{k = 1}^{\infty} \frac{t^{k}}{k!} \mathbf{A}^{k}.
      \]
      From the triangle inequality and the sub-multiplicative property of operator norms, we obtain
      \[
       \left\| \sum\limits_{k = 1}^{N} \frac{t^{k}}{k!} \mathbf{A}^{k} \right\|_{\infty} \leq \sum\limits_{k = 1}^{N} \frac{|t|^{k}}{k!} \|\mathbf{A}^{k} \|_{\infty} \leq \sum\limits_{k = 1}^{N} \frac{|t|^{k}}{k!} \|\mathbf{A}\|_{\infty}^{k}.
      \]
      Taking the limit when $N \to \infty$ on both sides we get the inequality
      \begin{equation}\label{eqn:exp_norm_bound}
      \|g_{t} - \Id \|_{\infty} \leq e^{|t|\| \mathbf{A} \|_{\infty}} - 1. 
      \end{equation}
      Noticing that $g_{t} \cdot x - x = (g_{t} - \Id) \cdot x$, and after a rearrangement of terms, it follows from equation (\ref{eqn:exp_norm_bound}) that choosing $t \in \R$ such that
        \begin{equation*}
            |t| \leq \frac{1}{\|\mathbf{A} \|_{\infty}}\log \left (\frac{\varepsilon}{\|x\|_{\infty}} + 1 \right),
        \end{equation*}
        ensures us that
        \begin{equation*}
            \|g_{t} \cdot x - x \|_{\infty} \leq \varepsilon,
        \end{equation*}
        as required.
  \end{proof}
  
  \begin{remark}
      The proof of Proposition \ref{prop:threshold} only makes use of general properties of vector norms and their induced operator norms, hence the result is also true by replacing the $\infty$-norm with any other norm in $\R^{n}$.
  \end{remark} 

      From Proposition \ref{prop:threshold} holding with complete generality, we will assume that the group elements satisfy a desired error threshold $\varepsilon$ for the remained of this work.

\section{Geometric Deep Learning}  GDL provides a formal framework to study the symmetry of deep learning models. Using the framework of GDL, we will study a certain symmetry group arising from condition 1) of Definition \ref{def:adv_attack_classical}; providing an algebraic framework of adversarial attacks. We first introduce GDL then discuss the symmetry group of interest. We then demonstrate how to compute the symmetry group explicitly and provide examples of three common deep learning architectures: Feed-forward multi-Layer perceptron (MLP), convolutional neural network (CNN) and graph convolutional network (GCN).

Let $\Omega$ be an input domain (images, graphs, etc.) and $G$ be a group acting on $\Omega$. Let $\Omega' \subseteq \Omega$. GDL provides a language to understand deep learning models as a composition of functions invariant under the symmetry group acting on the input domain. 

Under this framework, one can define any deep learning model as the following composition:
\begin{equation}
    F = A \circ \sigma_{J} \circ B_{J} \circ P_{J-1} \circ \ldots \circ P_{1} \circ \sigma_{1} \circ B_{1},
\end{equation}
where $A \colon \mathcal{X}(\Omega, \mathbb{F}) \to \mathcal{Y}$ is a $G$-invariant global pooling layer such that $A(g\cdot x) = A(x)$ for all $g \in G$ and $x \in \mathcal{X}(\Omega, \mathbb{F})$, $P \colon \mathcal{X}(\Omega, \mathbb{F}) \to \mathcal{X}(\Omega', \mathbb{F})$ is some local pooling layer, $\sigma$ is a non-linearity, and $B \colon \mathcal{X}(\Omega, \mathbb{F}) \to \mathcal{X}(\Omega', \mathbb{F}')$ is a linear $G$-equivariant layer such that $B(g \cdot x) = g \cdot B(x)$ for all $g \in G$ and $x \in \mathcal{X}(\Omega, \mathbb{F})$.

\subsection{Lie Groups of Neural Network Symmetries}
This subsection will deal with condition 2) of Definition \ref{def:alg_attack}. Recall that condition 2) of an adversarial attack $\tilde{x}$, to an input $x$ of a neural network $F$, states that one must have $F(\Tilde{x}) = F(x)$. Therefore, condition 2) can be phrased algebraically in terms of an element $g \in G$ such that $F(g \cdot x) = F(x)$. This leads us to study the groups acting on the input space of $F$, such that their induced actions leave $F$ invariant. 

\begin{definition}\label{def:neural_network_invariant}
    Let $F : X \to Y$ be a neural network and let $G$ be a group acting on $X$. We will say that $G$ is a symmetry group of $F$ if 
    \begin{equation*}
        g \cdot F = F,
    \end{equation*}
    for all $g$ in $G$.
\end{definition}

In terms of Definition \ref{def:neural_network_invariant}, we note that the choice of any element of a symmetry group will automatically satisfy condition 2). Indeed, if $g \in G$, $g^{-1}$ is also an element of $G$ and thus $(g^{-1} \cdot F)(x) = F(g \cdot x) = F(x)$.

In the following, we will always assume that $F : \R^{n} \to \R^{m}$ is a neural network of the form $F = f \circ L$, where the first layer $L : \R^{n} \to \R^{k}$ is an affine linear map, given by
\begin{equation}\label{def:first_layer}
    L(x) = Wx + b, \quad W \in \mathrm{M}_{k \times n}(\R), \; b \in \R^{k}.
\end{equation}
Under this assumption, we note that to find a symmetry group of $F$, it is sufficient to find a symmetry group of the first layer, $L$. Indeed, it follows from
    \[
    (g \cdot F)(x) = F(g^{-1} \cdot x) = f(L(g^{-1} \cdot x)) = f((g \cdot L)(x))
    \]
that any symmetry group of $L$ is also a symmetry group of $F$. We will proceed to analyse and discuss how to compute elements of the symmetry group of $L$. 

We proved in \cite{alg_attacks_icmlc} that the group $P_{W} \ltimes \ker(W)$, where
    \begin{equation*}
         P_{W} : = \lbrace g \in \GL_{n}(\R) \; : \; Wg = W \rbrace, 
    \end{equation*}
is a symmetry group of any neural network whose first layer is of the form given in equation (\ref{def:first_layer}). Clearly $P_{W} \ltimes \ker(W)$ is a subgroup of $\Aff_{n}(\R)$, and therefore a matrix Lie group. We remark that if $\rank W = n$, $P_{W}$ is the trivial group, namely $P_{W} = \lbrace \Id \rbrace$ (see \cite[Corollary 1]{alg_attacks_icmlc}). In other words, the number of neurons in $r$ must be less than the dimension of input $n$.

We let $\mathcal{W}$ denote the vector space spanned by the rows of $W$ for the remainder of this work. 

    We demonstrate in \cite{alg_attacks_icmlc} that one can generate elements of $P_{W}$ in terms of the Lie algebra $\lie{p}_{W}$, of $P_{W}$, via the correspondence between Lie groups and Lie algebras given by the exponential map defined in Section 3. We provide the results here for completeness.
    
    From Definition \ref{def:group_algebra}, the Lie algebra of a matrix Lie group $G \subseteq \GL_{n}(\R)$ is comprised by all $A \in \gl_{n}(\R)$ such that $\exp(tA) \in G$ for all $t \in \R$. Even more, the image of $\gl_{n}(\R)$ under $\exp$ is $\GL_{n}^{+}(\R)$. We define the group $P_{W}^{+} : = P_{W} \cap \GL_{n}^{+}(\R)$, which is equal to $\exp(\lie{p}_{W})$, and remark that $\lie{p}_{W}$ is the Lie algebra of both $P_{W}^{+}$ and $P_{W}$.

            \begin{proposition}\label{prop:lie_algebra}
                The Lie algebra of $P_{W}^{+}$ is 
                \[
                \lie{p}_{W} = \lbrace A \in \gl_{n}(\R) \; : \; Aw = 0, \; \forall w \in \mathcal{W} \rbrace.
                \]
            \end{proposition}
            \begin{proof}
            For any $A \in \lie{p}_{W}$, $\exp(tA) \in P_{W}$ for all $t \in \R$. Then, $\exp(tA) w = w$ for all $w \in \mathcal{W}$ and
            \[
            0 = \frac{d}{dt} \; \exp(tA)w = \exp(tA)Aw, \quad \forall t\in \R.
            \]
            Particularly, when $t = 0$, the above equation becomes $Aw = 0$, since $\exp(0) = \Id$.
            \end{proof}
            We will proceed to describe an algorithm to compute elements of $P_{W}^{+}$ explicitly, without the aid of any specific basis of $\R^{n}$. To this end, we define the linear map $A_{xy} \in \gl_{n}(\R)$, given by 
            \[
            A_{xy}z = xy^{t}z = \langle y, z \rangle x, \quad x, y, z \in \R^{n}.
            \]
            The kernel of $A_{xy}$ is precisely the vector space orthogonal to $\R y := \lbrace  k y \colon k \in \R \rbrace$. Noticing that $\mathcal{W}^{\perp} = \ker(W)$, it is not difficult to see from this construction that $\mathfrak{p}_{W} = \mathrm{span} \lbrace A_{xy} \; : \; y \in \ker(W) \rbrace$. The following proposition is a consequence of this construction and Proposition \ref{prop:lie_algebra}.
            \begin{theorem}\label{prop: PW_generated}
                $P^{+}_{W}$ is generated by 
                \[
                \left\lbrace \exp\left( \sum_{i = 1}^{k} A_{x_{i}y_{i}} \right) \; \colon \; y_{i} \in \ker(W), \; k \geq 1 \right\rbrace .
                \]
            \end{theorem}
            Theorem \ref{prop: PW_generated} thereby provides an explicit procedure to compute the elements of a symmetry group of any neural network whose first layer is an affine map.

            \begin{remark}
                In later sections, we take $G = P_{W} \cap \mathrm{O}_{n}$. The following proposition demonstrates how to generate elements of $P_{W} \cap \On_{n}$.
            \end{remark}
            \begin{proposition}
                Let $y \in \ker(W)$ and define the skew-symmetric matrix $\hat{A}_{xy} = \frac{1}{2}(xy^{t} - yx^{t})$. Then $\exp(\hat{A}_{xy}) \in P_{W} \cap \On_{n}$.  
            \end{proposition}
            \begin{proof}
            We want to show that $\exp(\hat{A}_{xy})^{t} = \exp(\hat{A}_{xy})^{-1}$. By construction $\hat{A}_{xy}^{t} = - \hat{A}_{xy}$, thus
            \begin{equation*}
                \exp(\hat{A}_{xy})\exp(\hat{A}_{xy})^{t} = \exp(\hat{A}_{xy})\exp(\hat{A}_{xy}^{t}) = \exp(\hat{A}_{xy}+ \hat{A}_{xy}^{t}) = \exp(0) = \Id.
            \end{equation*}
            So $\exp(\hat{A}_{xy}) \in \On(n)$, by construction $\exp(\hat{A}_{xy}) \in P_{W}$ and therefore, $\exp(\hat{A}_{xy}) \in P_{W} \cap \On_{n}$.
            \end{proof}

In the following sections we describe how to compute the symmetry group of three common deep learning architectures: Feed-forward MLPs, CNNs and GCNs.

\subsubsection{Feed-forward MLP}
Consider a $k$-layer MLP neural network $F : \R^{n} \to \R^{m}$, as the following composition of functions
        \begin{equation*}
            F = B_k \circ \sigma_{k-1} \circ B_{k} \circ \ldots \circ \sigma_{1} \circ B_{1}.
        \end{equation*}
        Here $\sigma_{i} \circ B_{i} : \R^{n_{i}} \to \R^{n_{i+1}}$, where $B_{i}(x) = \mathbf{W}_{i}x+b_{i}$ is an affine map with trained weight matrix $\mathbf{W}_{i} \in \mathbb{R}^{n_{i+1} \times n_{i}}$, $b_{i} \in \mathbb{R}^{n_{i+1}}$ is the bias, $\sigma_{i}$ is a non-linear activation function, and $n_{1} = n$, $n_{k} = m$. The pooling layers $P$ and $G$-invariant function are the identity maps.

    Following Section 4.1, to find a symmetry group of $F$, we note that to ensure $F(x) = F(g\cdot x)$ it is sufficient to find a symmetry group of the first layer, $B_{1}$. Since $B_{1}(x) = \mathbf{W}_{1}x + b_{1}$, it is of the form given by equation (\ref{def:first_layer}) and therefore $P_{\mathbf{W}_{1}} \ltimes \ker(\mathbf{W}_{1})$ is a symmetry group of $F$.

\subsubsection{Convolutional Neural Networks}
Following the geometric deep learning framework of Bronstein et al. \cite{bronstein2021geometricdeeplearninggrids} a convolutional layer can be described as the following composition 
\begin{equation}\label{def:convolutional_layer}
    h = P(\sigma(L(x)))
\end{equation}
where $L$ is a linear equivariant function. $P$ is a coarsening operation and $\sigma$ is a non-linearity. The symmetry group $G$ is translation where we realise a grey-scale image as a grid $\Omega = [H] \cross [W]$. The space of signals is $\mathcal{X}(\Omega, \R) = \lbrace x \colon \Omega \to \R \rbrace$ i.e. each pixel has a grey-scale value between $0$ and $1$. Given a compactly supported filter $f$ of size $H^{f} \cross W^{f}$, the convolution operation can be written as
\begin{equation*}
    L(x) = \sum^{H^{f}}_{v = 1}\sum^{W^{f}}_{w = 1} \alpha_{vw} \mathbf C(\theta_{vw}) x,
\end{equation*}
where $\mathbf{C}(\theta_{vw})$ is a circulant matrix with parameters $\theta_{vw}$. 

Let us assume that $F$ is CNN given by the composition
\begin{equation*}
    F = w \circ h_{1},
\end{equation*}
where each $h_{1}$ is a convolutional layer given by equation (\ref{def:convolutional_layer}) and $w$ is a composition of the other building blocks of a CNN (e.g. pooling layers and fully connected linear layers). Following our previous reasoning, in order to find a symmetry group $F$, it is sufficient to find a symmetry group for its first convolutional layer. In other words, we need to find a group $G$ such that $h_{1}(g \cdot x) = h_{1}(x)$, for all $g \in G$. Since $h_{1} = P_{1} \circ \sigma_{1} \circ L_{1}$, any symmetry group of $L_{1}$ is also a symmetry group of $h_{1}$ and thus of $F$. Noticing that $L_{1}$ is a linear map, it follows that $P_{L_{1}} \ltimes \ker(L_{1})$ is a symmetry group of $F$.

\subsubsection{Graph Convolutional Neural Networks}

Given a graph $G = (V,E)$ with nodal attributes $\mathbf{X} \in \R^{|V| \times f}$ where $f$ is the number of features. One can represent a graph with its adjacency matrix $\mathbf{A}$. Given a neighbourhood of a node $v \in V$ one has the associated multi-set of neighbourhood features
\begin{equation*}
    \mathbf{X}_{\mathcal{N}_{v}} = \{\{ x_{v} \colon u \in \mathcal{N}_{v}\}\}.
\end{equation*}
Given a permutation invariant function $\phi(x_v, \mathbf{X}_{\mathcal{N}_{v}})$ one defines a graph neural network $F(\mathbf{X}, \mathbf{A})$ as the matrix of $\phi$ applied to each node in the graph:

\begin{equation*}
F(\mathbf{X}, \mathbf{A}) = \begin{pmatrix} 
        \phi(x_{u_{1}}, \mathbf{X}_{\mathcal{N}_{u_{1}}})\\
    \vdots\\
     \phi(x_{u_{n}}, \mathbf{X}_{\mathcal{N}_{u_{n}}})
    \end{pmatrix}.
\end{equation*}

A graph convolution layer is defined as:
\begin{equation*}
    h_{v} = \phi(x_{v}, \bigoplus_{u \in \mathcal{N}_{v}} c_{uv}\psi(x_{u})),
\end{equation*}
where, $\psi(x) = \mathbf{W}x +b$, $\phi(x,z) = \sigma(\mathbf{W}x+\mathbf{U}z+b)$ and $c_{uv} \in \R$. Noticing that $\psi$ is an affine transformation, to find the symmetry group it suffices to find the symmetry group of $\psi$, the same group as the feed-forward MLP. Since $\psi(x) = \mathbf{W}x + b$ is of the form given by equation (\ref{def:first_layer}), $P_{\mathbf{W}} \ltimes \ker(\mathbf{W})$ is, therefore, a symmetry group of $h_v$.

\begin{remark}
    As noted by Bronstein et al. \cite{bronstein2021geometricdeeplearninggrids} other graph neural network architectures (e.g. message passing and attention) have a similar formulation as above and therefore one can compute similar symmetry groups.
\end{remark}

\section{Algebraic Adversarial Attacks on Explainablity Models}
In this section we propose algebraic adversarial examples for explainability models. We demonstrate through exploiting the symmetry of a neural network alone, one can produce adversarial examples. Our approach removes the need for an optimisation scheme to generate adversarial examples. We apply our theory to three explainability models: Path-based attribution methods, neural conductance and LIME. Each subsection demonstrates that elements of the symmetry group $P_{W} \ltimes \ker(W)$ are algebraic adversarial attacks on the aforementioned explainability models.

  \subsection{Equivariant Path Methods}
  \label{subsec:equiv_path_methods}

  Base-point attribution methods (BAM) \cite{lundstrom2022rigorous} are a specific class of post hoc explainability models. A BAM is a function 
\begin{align*}
 A \colon &\R^{n} \times \R^{n} \times F(\R^{n}, \R^{m}) \to \mathbb{R}^{n}\\
 &(x,x',F) \mapsto A(x,x',F)
\end{align*}
 where, $F(\R^{n}, \R^{m})$ is the vector space of functions from $\R^{n}$ to $\R^{m}$ and $x,x' \in \R^{n}$ are an input and a base-point, respectively. 
Path methods are a specific class of base-point attribution method. Given a closed interval $I := [a, b] \subset \R$, a path $\gamma \colon I \to \R^{n}$ and a unit length vector $v \in \R^{n}$, the component of a path method $A^{\gamma} : \R^{n} \times \R^{n} \times F(\R^{n},\R^{m}) \to \R^{n}$ in the direction of $v$ is defined as
\begin{equation}\label{def:BAM}
    A^{\gamma}_{v}(x,x',F) = \int_{a}^{b} \langle \nabla F(\gamma(t)), v \rangle \langle \gamma'(t), v \rangle \dd t.
\end{equation}
In this way, for an orthonormal basis $\lbrace v_{1}, \dots, v_{n} \rbrace$ of $\R^{n}$, $A^{\gamma}$ is given by
\begin{equation}\label{eqn:attribution_basis}
    A^{\gamma}(x,x',F) = ( A^{\gamma}_{v_{1}}(x,x',F), \ldots, A^{\gamma}_{v_{n}}(x,x',F)).
\end{equation}
Particularly, for the standard orthonormal basis $\lbrace e_{1}, \dots, e_{n} \rbrace$ of $\R^{n}$, we obtain the usual definition
\begin{equation}
    A^{\gamma}_{e_{i}}(x,x',F) = \int_{a}^{b} \frac{\partial F}{\partial x_{i}}(\gamma(t)) \frac{\partial \gamma_{i}}{\partial t}(t) \dd t.
\end{equation}

The prominent path method, integrated gradients \cite{pmlr-v70-sundararajan17a} is a path method where $\gamma$ is taken to be the straight line between points $x,x' \in \R^{n}$. For any pair of points $x, x' \in \R^{n}$, a neural network $F \in F(\R^{n},\R^{m})$, and the standard orthonormal basis of $\R^{n}$ integrated gradients is defined as:
\[
\mathrm{IG}(x, x', F) := (x - x') \odot \int_{0}^{1} \nabla F(x' + t(x - x')) \dd t,
\]
where $\odot$ denotes the Hadamard product.

Integrated gradients has the particularity that all of its paths behave well under the action of the group of Euclidean transformations in the following sense: Consider the path $\gamma_{x'}^{x}(t) = x' + t (x - x')$, from $x'$ to $x$, and $g \in \mathrm{E}_{n}$ \cite{alg_attacks_icmlc}. Acting with $g$ on $x$ and $x'$, we can easily see that the straight line from $g \cdot x'$ to $g \cdot x$ is nothing but $g \cdot \gamma_{x'}^{x}(t)$. This motivates the following definition:
\begin{definition}
    Let $\gamma_{x'}^{x}$ be a path in $\R^{n}$ from $x'$ to $x$ and let $G$ be a group acting on $\R^{n}$. We say that $\gamma_{x'}^{x}$ is $G$-equivariant if
    \begin{equation*}
        \gamma_{g \cdot x'}^{g \cdot x}(t) = g \cdot \gamma_{x'}^{x}(t),
    \end{equation*}
    for all $g \in G$. 
\end{definition}

    Integrated gradients is a path method whose curves are straight lines, hence $\Aff_{n}(\R)$-equivariant. This condition will become essential for the remainder of this section. It was proved in \cite[Proposition 8]{alg_attacks_icmlc} that the $\mathrm{E}_{n}$-equivariance of straight lines implies the $E_{n}$-invariance of integrated gradients in the following sense:  
    
    \begin{definition}
        Let $A : \R^{n} \times \R^{n} \times F(\R^{n}, \R) \to \R^{n}$ be a BAM and $G$ be a group acting on $\R^{n}$. We will say that $A$ is $G$-invariant if 
        \[
        A_{g \cdot v}(g\cdot x, g \cdot x', g \cdot F) = A_{v}(x, x', F),
        \]
        for all $x, x' \in \R^{n}$, $F \in F(\R^{n}, \R)$ and $v \in \R^{n}$ of unit length.
    \end{definition}  

    \begin{remark}
        Not all group actions in $\R^{n}$ preserve the length of vectors. In the cases when $G$ is a subgroup of $\Aff_{n}(\R)$, $A_{g \cdot v}(x, x', F)$ will mean that only the $\mathrm{O}_{n}$ component of $g$ is acting on $v$. In the following we will let $S^{n-1} : = \lbrace v \in \R^{n} \; : \; \| v \|_{2} = 1 \rbrace$ be the $(n-1)$-dimensional sphere in $\R^{n}$.
    \end{remark}
  
  Below, \cite[Proposition 8]{alg_attacks_icmlc} is generalised to path methods defined by $\Aff_{n}(\R)$-equivariant curves.

  \begin{theorem}\label{theo:aff_invariance}
      Let $G$ be a subgroup of $\Aff_{n}(\R)$, $\gamma$ a $G$-equivariant path from $x'$ to $x$, $A^{\gamma}$ a path method and $v \in S^{n-1}$. Then
        \begin{equation}\label{ig_on_sym1}
            A^{\gamma}_{Qv}(Qx, Qx', Q \cdot F) = A^{\gamma}_{v}(x, x', F), \quad \forall Q \in \On_{n} \cap G,
        \end{equation}
        \begin{equation}\label{ig_t_sym3}
            A^{\gamma}_{v}(x + u, x' + u, u \cdot F) = A^{\gamma}_{v}(x, x', F), \quad \forall u \in \R^{n} \cap G,
        \end{equation}
        \begin{equation}\label{ig_sym_sym2}
            A^{\gamma}_{v}(\exp(S) x, \exp(S)x', \exp(S) \cdot F) = A^{\gamma}_{v}(x, x', F),
        \end{equation}
        for all $S \in \Sym_{n}(\R)$ such that $\exp(S) \in G$ and $v$ is an eigenvector of $S$. Furthermore, $A^{\gamma}$ is a $G$-invariant path method, i.e.
        \begin{equation}\label{ig_on_sym2}
            A^{\gamma}_{g \cdot v}(g \cdot x, g \cdot x', g \cdot F) = A^{\gamma}_{v}(x, x', F),
        \end{equation}
        for all $g \in G$.
  \end{theorem}

  \begin{proof}
    For the proofs of equations (\ref{ig_on_sym1}) and (\ref{ig_t_sym3}) see \cite[Proposition 8]{alg_attacks_icmlc}. To verify that equation (\ref{ig_sym_sym2}) holds, recall that if $Sv = \lambda v$ then $\exp(S)v = e^{\lambda}v$, and $\exp(S) \in \GL_{n}(\R)$. By the chain rule, we note that
    \[
    \nabla \exp(S) \cdot F = \exp(S) \cdot (\exp(S^{t})^{-1}\nabla F) = \exp(S) \cdot (\exp(-S) \nabla F),
    \]
    where the last equality follows from the fact that $S$ is symmetric and that $\exp(S)^{-1} = \exp(-S)$. Under the assumption that $\gamma$ is $G$-equivariant, we observe that
    \[
    \langle \nabla \exp(S) \cdot F (\gamma_{\exp(S)x'}^{\exp(S)x}(t)), v \rangle = \langle (\exp(S) \cdot (\exp(-S)\nabla F )(\exp(S) \gamma_{x'}^{x}(t)), v \rangle = \langle \nabla F(\gamma_{x'}^{x}(t)), \exp(-S)v \rangle.
    \]
    Since $S v = \lambda v$, we obtain
    \begin{equation}\label{eqn:dilation1}
    \langle \nabla \exp(S) \cdot F (\gamma_{\exp(S)x'}^{\exp(S)x}(t)), v \rangle = e^{-\lambda} \langle \nabla F(\gamma_{x'}^{x}(t)), v \rangle.
    \end{equation}
    On the other hand
    \begin{equation}\label{eqn:dilation2}
    \langle (\gamma_{\exp(S)x'}^{\exp(S)x})'(t), v \rangle  = \langle \gamma_{x'}^{x}(t), \exp(S)v \rangle = e^{\lambda} \langle \gamma_{x'}^{x}(t), v \rangle
    \end{equation}
    follows again from the equivariance of $\gamma$ and the fact that $v$ is an eigenvector of $S$. Combining equations (\ref{eqn:dilation1}) and (\ref{eqn:dilation2}), we get
    \[
    A_{v}^{\gamma}(\exp(S) x, \exp(S)x', \exp(S) \cdot F) = \int_{a}^{b} \langle \nabla F(\gamma_{x'}^{x}(t)), v \rangle \langle \gamma_{x'}^{x}(t), v \rangle \dd t,
    \] 
    as claimed.

    Lastly, let $g \in G$ be of the form $g = (Q\exp(S), u)$, where $Q\exp(S)$ is the polar decomposition of an element of $\GL_{m}(\R)$ (see Proposition \ref{prop:polar_decomp} ). It follows from equations (\ref{ig_on_sym1}), (\ref{ig_t_sym3}) and $(\ref{ig_sym_sym2})$ that
    \begin{align*}
      A^{\gamma}_{g \cdot v}(g \cdot x, g \cdot x', g \cdot F) &= A^{\gamma}_{Qv}(Q\exp(S)x + u, Q\exp(S)x' + u, (Q\exp(S), u) \cdot F) \\
     & = A^{\gamma}_{Qv}(Q\exp(S)x, Q\exp(S)x', Q\exp(S) \cdot F) \\
     & = A^{\gamma}_{v}(\exp(S)x, \exp(S)x', \exp(S) \cdot F) \\
     & = A^{\gamma}_{v}(x, x', F),   
    \end{align*}
    as required.
  \end{proof}
    The above proposition provides a generalisation of \cite[Proposition 8]{alg_attacks_icmlc} and motivates our definition of invariance by affine transformations of path methods. The literature only considers the invariance of path methods under the action of $n$ copies of $\Aff_{1}(\R)$ on $\R^{n}$ \cite{lundstrom2022rigorous,pmlr-v70-sundararajan17a,  Friedman99} and not the whole group $\Aff_{n}(\R)$. Under our framework, it corresponds with the group $\Aff_{1}(\R) \times \dots \times \Aff_{1}(\R)$, as a subgroup of $\Aff_{n}(\R)$, with the following action:
    \[
    ((a_{1}, b_{1}), \dots, (a_{n}, b_{n})) \cdot x = (a_{1}x_{1} + b_{1}, \dots, a_{n}x_{n} + b_{n}),
    \]
    where $a_{i} \neq 0$ and $x = (x_{1}, \dots, x_{n}) \in \R^{n}$. We refer to \cite[Section 3.3]{lundstrom2022rigorous} for further discussion.

    In \cite{alg_attacks_icmlc} we provide two examples of adversarial attacks for integrated gradients. Here we extend these attacks to base-line attribution methods invariant by affine transformations.

    \begin{theorem}
    \label{prop:mult_ig_attack}
        Let $F : \R^{n} \to \R$ be a neural network whose first layer is $L(x) = Wx + b$, and let $A$ be a $G$-invariant base-line attribution method, for some $G \leq \Aff_{n}(\R)$. Suppose that the set
        \[
        \lbrace (v, x') \in S^{n-1} \times \R^{n} \; : \; A_{g^{-1} \cdot v}(x, g^{-1} \cdot x', F) = A_{v}(x, x', F)\rbrace
        \]
        has measure 0. Then $\Tilde{x} = g \cdot x$ is an algebraic adversarial attack to $(A, F, x)$ for all $g \in P_{W} \cap \On_{n}\ltimes \ker(W)$, such that $g \neq (\Id, 0)$.
    \end{theorem}
    \begin{proof}
        By Proposition \ref{prop:threshold}, \cite[Proposition 1]{alg_attacks_icmlc} and \cite[Proposition 2]{alg_attacks_icmlc}, conditions 1 and 2 are satisfied. It is only left to verify that $\tilde{x} = g \cdot x$ satisfies condition 3. By Proposition \label{theo:invariance}, for any $g \in P_{W} \cap \On_{n}\ltimes \ker(W)$, we have that
        \[
        A_{v}(g\cdot x, x', F) = A_{v}(g \cdot x, gg^{-1} \cdot x', gg^{-1}  \cdot F) = A_{g^{-1} \cdot v}(x, g^{-1} \cdot x', F).
        \]
        By assumption, the set $\lbrace (v, x') \in S^{n-1} \times \R^{n} \; : \; A_{g^{-1} \cdot v}(x, g^{-1} \cdot x', F) = A_{v}(x, x', F)\rbrace$ has measure 0, which guarantees that $A_{g^{-1} \cdot v}(x, g^{-1} \cdot x', F) \neq A_{v}(x, x', F)$ with probability 1. Consequently, $A(g \cdot x, x', F) \neq A(x, x', F)$ as claimed.
    \end{proof}

    \begin{remark}
        In practice, the assumption of the set $\lbrace (v, x') \in S^{n-1} \times \R^{n} \; : \; A_{g^{-1} \cdot v}(x, g^{-1} \cdot x', F) = A_{v}(x, x', F)\rbrace$ having measure 0 will hold. This set is the preimage of the point $A_{v}(x, x', F)$ by the map $f: S^{n-1} \times \R^{n} \to \R$, $(v, x') \mapsto A_{g^{-1} \cdot v}(x, g^{-1} \cdot x', F)$. $S^{n-1} \times \R^{n}$ has dimension $2n - 1$ and generally $f^{-1}(\lbrace A_{v}(x, x', F)\rbrace)$ will have dimension $\leq 2n-2$.
    \end{remark}
        
    Regarding Integrated Gradients, the corollary below follows immediately from Theorem \ref{theo:aff_invariance} and Theorem \ref{prop:mult_ig_attack}

    \begin{corollary}
        Let $F : \R^{n} \to \R$ be a neural network whose first layer is $L(x) = Wx + b$. Then $\Tilde{x} = g \cdot x$ is an algebraic adversarial attack to $(\mathrm{IG}, F, x)$ for all $g \in P_{W}\ltimes \ker(W)$, such that $g \neq \Id$.
    \end{corollary}
 
The condition $\|g\cdot x - x\|_{\infty} \leq \varepsilon$ restricts the extent to which the adversarial and clean point may differ. The following proposition demonstrates that the error threshold bounds the difference between the clean and adversarial explanations. 

\begin{proposition}
\label{prop:ig_error}
    Let $g \in \Aff_{n}(\R)$ be an algebraic attack on integrated gradients such that $\|g \cdot x - x \|_{\infty} \leq \varepsilon$. Suppose $F$ is $L$-Lipschitz. The difference between the adversarial and clean explanation for integrated gradients in the direction of unit vector $v$ has the following bound:
    \begin{equation*}
    | \IG_{v}(x,x',F) - \IG_{v}(g \cdot x, x',F) | \leq (2\| x - x' \|_{2} + \varepsilon \sqrt{n}) L.
    \end{equation*} 
\end{proposition}

\begin{proof}
    Let $\tilde{x} = g \cdot x$ for $g \in \Aff_{n}(\R)$ such that $\tilde{x} = g \cdot x$ is an algebraic adversarial attack with $\|g \cdot x - x \| \leq \varepsilon$. Applying the triangle and Cauchy-Schwartz inequalities provides the following inequality:

    \begin{align*}
     |\IG_{v}(x,x',F) - \IG_{v}(\tilde{x},x' ,F) | &\leq \int_{0}^{1} (\|\nabla F(\gamma_{x'}^{x}(t))\|_{2} \| (\gamma_{x'}^{x})'(t)\|_{2} + \|\nabla F(\gamma_{x'}^{\Tilde{x}}(t)) \|_{2} \|(\gamma_{x'}^{\Tilde{x}})'(t)\|_{2}) \|v\|^{2}_{2} \dd t.
     \end{align*}
     Since the Lipschitz constant of $F$ is $L$, $\| v \|_{2} = 1$ and $(\gamma_{x'}^{x})'(t) = x - x'$, we get
     \begin{align*}
     |\IG_{v}(x,x',F) - \IG_{v}(\tilde{x},x' ,F) | &  \leq  \int_{0}^{1} L \|x - x'\|_{2} + L \|_{2} (\|g \cdot x - x + x - x' \|_{2}) \; \dd t\\
     & \leq  L \int_{0}^{1} \|x - x'\|_{2} + (\|x - x'\|_{2} + \|g \cdot x - x \|_{2}) \; \dd t\\
     &\leq L \int_{0}^{1}  2 \|x - x' \|_{2} + \varepsilon \sqrt{n} \; \dd t\\
     & = (2 \|x - x'\|_{2} + \varepsilon \sqrt{n}) L.
    \end{align*}
\end{proof}

    \subsection{Neural Conductance}
    Path methods quantify the importance of the input features for a neural networks classification. Neural conductance extends path methods to determine the importance of a neuron in a network. Here we extend algebraic adversarial attacks to path methods.

    Let $A^{\gamma} : \R^{n} \times \R^{n} \times F(\R^{n}, \R) \to \R^{n}$ be a path method for some curve $\gamma : [0, 1] \to \R^{n}$ such that $\gamma(0) = x'$ and $\gamma(1) = x$. Given a neural network $F \in F(\R^{n}, \R)$, one can decompose $F(x) = f(h(x))$ where $h \colon \R^{n} \to \R^{m}$ is the output of the neural network up to some layer and $f \colon \R^{m} \to \R$ is the second half of the neural network taking the output of the hidden layer $h$ as the input. Following \cite{Neural_conductance_original}, the flow of the gradient in $A^{\gamma}_{i}(x, x', F)$ through a neuron $j$ is defined as:
    \begin{equation*}
        A^{\gamma}_{i,j}(x, x', F) \coloneqq (x_{i} - x_{i}')\int^{1}_{0} \frac{\partial f}{\partial h_{j}} (h(\gamma(t))) \frac{\partial h_{j}}{\partial x_{i}}(\gamma(t)) \dd t.
    \end{equation*}

    The cumulative impact of a neuron $j$ is called neural conductance and is defined as:
    \begin{equation}\label{IG_neural_conductance}
        A^{\gamma}_{*,j}(x, x', F) \coloneqq \sum\limits_{i = 1}^{n} A^{\gamma}_{i,j}(x, x', F) =  \int_{0}^{1} \frac{\partial f}{\partial h_{j}} (h(\gamma(t))) \frac{\dd (h_{j} \circ \gamma)}{\dd t}(t) \dd t.
    \end{equation}
    
    Here we provide a generalisation of neural conductance of arbitrary path methods within our framework. For any pair $v \in \R^{n}$, $w \in \R^{m}$ of vectors of unit length, we define the flow of $A_{v}^{\gamma}(x, x', F)$ in the direction of $w$ as
    \[
    A^{\gamma}_{v, w}(x, x', F) : = \int_{0}^{1} \langle \nabla f (h(\gamma(t))), w \rangle \langle \nabla h_{w} (\gamma(t)), v \rangle \langle \gamma'(t), v \rangle \dd t,
    \]
    where $h_{w}$ denotes the component of $h$ in the direction of $w$, namely $h_{w}(x) = \langle h(x), w \rangle$. Clearly, by replacing $v$ and $w$ with vectors from the standard bases of $\R^{n}$ and $\R^{m}$, respectively, we recover the classical definition of the flow through a neuron. In this description, $w$ is a linear combination of neurons such that its norm is equal to 1. Moreover, choosing any orthonormal basis $\lbrace w_{1}, \dots, w_{m} \rbrace$ of $\R^{m}$, by making use of the chain rule, we can recover $A^{\gamma}_{v}(x, x', F)$ by
    \[
    A^{\gamma}_{v}(x, x', F) = \sum\limits_{i = 1}^{m} A^{\gamma}_{v, w_{i}}(x, x', F).
    \]
    Analogously to equation (\ref{IG_neural_conductance}), we denote the neural conductance in the direction of $w$ by $A^{\gamma}_{*, w}$, which is explicitly given by
    \begin{equation}\label{path_neural_conductance}
        A^{\gamma}_{*, w}(x, x', F) := \sum\limits_{i=1}^{n} A^{\gamma}_{v_{i}, w}(x, x', F)
    \end{equation}
    for any orthonormal basis $\lbrace v_{1}, \dots, v_{n} \rbrace$ of $\R^{n}$.
    \begin{proposition}\label{prop:neural_conductance}
        Let $A^{\gamma} : \R^{n} \times \R^{n} \times F(\R^{n}, \R) \to \R^{n}$ be a path method defined by a curve $\gamma : [0, 1] \to \R^{n}$ such that $\gamma(0) = x'$ and $\gamma(1) = x$, and let $F = f \circ h$ with $f : \R^{m} \to \R$ and $h : \R^{n} \to \R^{m}$. Then
    \begin{equation}\label{new_neural_conductance}
    A^{\gamma}_{*, w}(x, x', F) = A_{w}^{h \circ \gamma}(h(x), h(x'), f),
    \end{equation}
    for all $w \in \R^{m}$ such that $\| w \|_{2} = 1$.
    \end{proposition}
    \begin{proof}
        Let $\lbrace v_{1}, \dots, v_{n} \rbrace$ be an orthonormal basis of $\R^{n}$. By definition, $A^{\gamma}_{*, w}(x, x', F)$ is given by
        \begin{equation}\label{eqn:conductance1}
        \sum\limits_{i=1}^{n} A^{\gamma}_{v_{i}, w}(x, x', F) = \int_{0}^{1} \langle \nabla f (h(\gamma(t))), w \rangle 
        \sum\limits_{i = 1}^{n}
        \langle \nabla h_{w} (\gamma(t)), v_{i} \rangle \langle \gamma'(t), v_{i} \rangle \dd t.
        \end{equation}
        Since $h \circ \gamma$ is a curve in $\R^{m}$, we note that $\langle (h \circ \gamma)(t), w \rangle = (h_{w} \circ \gamma) (t)$, hence
        \begin{equation}\label{eqn:conductance2}
        \langle (h \circ \gamma)'(t), w \rangle = \frac{\dd (h_{w} \circ \gamma)}{\dd t}(t) = \sum\limits_{i = 1}^{n}
        \langle \nabla h_{w} (\gamma(t)), v_{i} \rangle \langle \gamma'(t), v_{i} \rangle.
        \end{equation}
        Substituting equation (\ref{eqn:conductance2}) in equation (\ref{eqn:conductance1}) we get the desired result
        \[
        A^{\gamma}_{*, w}(x, x', F) = \int_{0}^{1} \langle \nabla f (h(\gamma(t))), w \rangle \langle (h \circ \gamma)'(t), w \rangle \dd t = A^{h \circ \gamma}_{w} (h(x), h(x'), f),
        \]
        as $h \circ \gamma$ is a path from $h(x')$ to $h(x)$ in $\R^{m}$.

    \end{proof}
     Lundstrom et al. \cite{lundstrom2022rigorous} note that neural conductance $\IG_{*,j}$ is a base-line path method since $h \circ \gamma$ is a path itself. The analogous statement for any path method is also true in our construction, as a consequence of Proposition \ref{prop:neural_conductance}. 
    
    To apply the results from Section \ref{subsec:equiv_path_methods} to neural conductance we require that the decomposition of $F$ satisfies the following conditions:
     \begin{enumerate}
         \item[1.] $h \circ \gamma$ is a $G$-equivariant path with respect to a subgroup $G$ of $\Aff_{m}(\R)$.
         \item[2.] The first layer of $f$ is an affine linear map.
     \end{enumerate}
    Condition 1) guarantees us, by Theorem \ref{theo:aff_invariance}, that $A^{h \circ \gamma}$ will be $G$-invariant. In addition, condition 2) is required to be able to compute a symmetry group for $f$, as described in Section 5. Therefore, Theorem \ref{prop:mult_ig_attack} provide us with examples of algebraic adversarial attacks to path methods via neural conductance. 
    
    Lastly, we remark that finding a decomposition fulfilling the above requirements is a feasible task for the types of neural networks described in Section 5, namely feed-forward MLPs, CNNs, and GCNs, as each of these neural networks is composed of layers containing affine linear maps.

\subsection{Smooth Grad and LIME}
In this subsection we provide a generalisation of Smooth Grad and prove analogous results for Smooth Grad as those obtained for path methods and neural conductance in Section \ref{subsec:equiv_path_methods}. 

Smooth Grad with $d$ samples and covariance matrix $\Sigma$ is the attribution method $\mathrm{SG} : F(\R^{n}, \R) \times \R^{n} \times \Sym_{n}(\R) \to \R^{n}$ defined by
 \begin{equation*}
    \mathrm{SG}(F, x, \Sigma) = \frac{1}{d}\sum_{i = 1}^{d} \nabla F(x + a_{i}),
\end{equation*}
where $F \in F(\R^{n}, \R)$, $x \in \R^{n}$, $\Sigma \in \Sym_{n}(\R)$ is a covariance matrix and $a_{i} \sim \mathcal{N}(0,\Sigma)$. The $i$-th component of smooth grad corresponds to the attribution to the $i$-th variable of $F$. Here we propose, analogously to integrated gradients, to define the smooth grad attribution in the direction of a unit vector $v \in S^{n-1}$ as:
\begin{equation}
        \mathrm{SG}_{v}(F, x, \Sigma) = \frac{1}{d}\sum_{i = 1}^{d} \langle \nabla F(x + a_{i}), v \rangle.
\end{equation}
Therefore, given an orthonormal basis $\lbrace v_{1}, \dots, v_{n} \rbrace$ of $\R^{n}$, we can express smooth grad by
\[
\mathrm{SG}(F, x, \Sigma) = (\mathrm{SG}_{v_{1}}(F, x, \Sigma), \ldots, \mathrm{SG}_{v_{n}}(F, x, \Sigma)).
\]
We remark that by choosing the usual orthonormal basis of $\R^{n}$, we recover the usual definition of Smooth Grad.

In the following proposition, for a matrix $A \in \GL_{n}(\R)$, we will write $A^{-t} = (A^{t})^{-1}$, and the action of $\Aff_{n}(\R)$ on $\Sym_{n}(\R)$ will be given by $(A, u) \cdot M = AMA^{t}$, where $M \in \Sym_{n}(\R)$.
\begin{proposition}\label{theo:SG_symmetries}
    Let $F : \R^{n} \to \R$ be a neural network and $v \in S^{n-1}$. Then 
    \begin{equation}
        \mathrm{SG}_{g \cdot v}(g \cdot F, g \cdot x, g \cdot \Sigma) = \mathrm{SG}_{v}(F, x, \Sigma),
    \end{equation}
    for all $g \in \Aff_{n}(\R)$. 
\end{proposition}

\begin{proof}
 When we pick $g \in \Aff_{n}(\R)$ to be a translation, i.e. $g = (\Id, u)$, the proof follows from a direct calculation. Let $g = (A, 0) \in \Aff_{n}(\R)$. A direct application of the chain rule shows that
 \begin{equation}\label{eqn:chain_rule_action}
 \nabla (g \cdot F)(x) = A^{-t}\nabla F(A^{-1}x).
 \end{equation}
 Choosing a unit length vector $v$, it is not difficult to see that
\begin{equation}\label{eqn:SG_1}
\mathrm{SG}_{g \cdot v}(g \cdot F, g \cdot x, g \cdot \Sigma) = \frac{1}{d}\sum\limits_{i=1}^{d} \langle \nabla(g \cdot F)(Ax + a_{i}), Av \rangle = \frac{1}{d}\sum\limits_{i=1}^{d} \langle A^{-t}\nabla F(x + A^{-1}a_{i}), Av \rangle
\end{equation}
follows from equation (\ref{eqn:chain_rule_action}) and $a_{i} \sim \mathcal{N}(0, g \cdot \Sigma) = \mathcal{N}(0, A\Sigma A^{t})$. A direct calculation shows that since $a_{i} \sim \mathcal{N}(0, A\Sigma A^{t})$, then $A^{-1}a_{i} \sim \mathcal{N}(0, A^{-1}A\Sigma A^{t}A^{-t}) = \mathcal{N}(0, \Sigma)$. Consequently, the right-hand side of equation (\ref{eqn:SG_1}) is nothing but $\mathrm{SG}_{v}(F, x, \Sigma)$.
\end{proof}

Exploiting the symmetries of Smooth Grad, we provide algebraic adversarial attacks, in an analogous way as for Integrated Gradients.

\begin{theorem}\label{prop:SG_attack}
    Let $F : \R^{n} \to \R$ be a neural network whose first layer is $L(x) = Wx + b$. Then $\Tilde{x} = g \cdot x$ is an algebraic adversarial attack to $(\mathrm{SG}, F, x, \Sigma)$ for all $g \in P_{W}\ltimes \ker(W)$, such that $g \neq (\Id, 0)$.
\end{theorem}
\begin{proof}
    It follows by applying Proposition \ref{theo:SG_symmetries} and following the proof of Theorem \ref{prop:mult_ig_attack}.
\end{proof}

In \cite{agarwal2021unification} Agarwal et al. prove the following equivalence between Smooth Grad and C-LIME, an extension of LIME to a continuous input space instead of binary variables.

\begin{theorem}\cite[Theorem 1]{agarwal2021unification}.
\label{thrm:lime_sg_equiv}
    Let $F \colon \R^n \to \R$ be a function. Then, for any $x \in \R^n$ and invertible covariance matrix $\Sigma \in \GL_{n}(\R)$,
    \begin{equation*}
        \mathrm{SG}(F, x, \Sigma) = \mathrm{LIME}_{\Sigma}^{F}(x).
    \end{equation*}
\end{theorem}
Agarwal et al's theorem implies that an adversarial attack on Smooth Grad is also an adversarial attack on C-LIME. Consequently, this result combined with Theorem \ref{prop:SG_attack}, directly imply the following result.

\begin{theorem}
\label{thrm:alg_attack_lime}
    Let $\tilde{x} = g \cdot x$ be an algebraic adversarial attack to $(\mathrm{SG}, F, x, \Sigma)$. Then $\tilde{x} = g \cdot x$ is an algebraic adversarial attack on $(\mathrm{LIME}, F, x, \Sigma)$.
\end{theorem}

\begin{proof}
  By Theorem \ref{prop:SG_attack} $\Tilde{x}$ is an algebraic attack on $(\mathrm{SG}, F, x, \Sigma)$. Therefore, by Theorem \ref{thrm:lime_sg_equiv} $\Tilde{x}$ is an algebraic attack on $(\mathrm{LIME}, F, x, \Sigma)$.
\end{proof}

\begin{proposition}
    \label{prop:sg_error_bound}
    Let $F : \R^{n} \to \R$ be a Lipschitz neural network with Lipschitz constant $L \in \R$, $x \in \R^{n}$ and $g \in \Aff_{n}(\R)$ such that $\| x - g \cdot x \|_{\infty} \leq \varepsilon$. Then
    \[
    | \mathrm{SG}_{v}(F, x, \Sigma) - \mathrm{SG}_{v}(F, g \cdot x, \Sigma)| \leq \varepsilon L \sqrt{n}.
    \]
\end{proposition}
\begin{proof}
    From the definition of Smooth Grad and the triangle and Cauchy-Schwarz inequalities we have  
    \[
    | \mathrm{SG}_{v}(F, x, \Sigma) - \mathrm{SG}_{v}(F, g \cdot x, \Sigma)| \leq \frac{1}{d} \sum\limits_{i = 1}^{d} \| \nabla F(x + a_{i}) - \nabla F(g \cdot x + a_{i}) \|_{2} \| v \|_{2}.
    \]
    Since $F$ is Lipschitz and $\| v \|_{2} = 1$, we note that the above inequality becomes
    \[
    \begin{array}{rcl}
     | \mathrm{SG}_{v}(F, x, \Sigma) - \mathrm{SG}_{v}(F, g \cdot x, \Sigma)| & \leq & \frac{1}{d} \sum\limits_{i = 1}^{d} L \| x + a_{i} - g\cdot x - a_{i} \|_{2} \\ 
     & \leq & L \| x - g \cdot x \|_{2}  \\
     & \leq & L \sqrt{n} \| x - g \cdot x \|_{\infty} \\
     & \leq & \varepsilon L \sqrt{n},
    \end{array}
    \]
    where the last inequality follows from the assumptions.
\end{proof}

\begin{corollary}
Let $F : \R^n \to \R$ be a Lipschitz neural network with Lipschitz constant $L \in \R$, $x \in \R^n$ and $g \in \Aff_{n}(\R)$ such that $\|g\cdot x - x\|_{\infty} \leq \varepsilon$. Then
\begin{equation*}
    |\mathrm{LIME}(F, x, \Sigma) - \mathrm{LIME}(F, g\cdot x, \Sigma)| \leq \varepsilon L \sqrt{n}.
\end{equation*}
\end{corollary}
\begin{proof}
    The result follows from Theorem \ref{thrm:alg_attack_lime} and Proposition \ref{prop:sg_error_bound}.
\end{proof}

\section{Experiments}
\subsection{Experimental Setup}
\begin{table}[t]
\centering
\caption{Train and test classification performance of MLP applied to: MNIST, Fashion-MNIST, Wisconsin breast cancer and network traffic datasets. The dimensions of input $n$, and number of neurons $r$ in the first layer of the MLP is provided for each dataset.}
\label{table:mlp_performance}
\begin{tabular}[]{lcccc}
\toprule
Dataset & Train Accuracy & Test Accuracy &  Train F1 & Test F1\\ \midrule
MNIST $(n = 784, r = 16)$ & 0.9826 &  0.9606 & 0.9824 & 0.9602\\
Wisconsin Breast Cancer $(n = 30, r = 16)$ &0.9685 &  0.9681 & 0.9663 & 0.9656\\
Network Traffic $(n = 20, r = 8)$ &0.9720 &  0.9727 & 0.9683 & 0.9691\\
\bottomrule
\end{tabular}
\end{table}%
We validate our approach of algebraic adversarial examples on two well-known classification datasets: MNIST \cite{mnist} and Wisconsin Breast Cancer \cite{misc_breast_cancer_wisconsin_(diagnostic)_17}, and one real-world example: Network traffic of Mobile Devices \cite{simpson2023testbed} (for each dataset a 2-layer MLP is trained for classification). Integrated gradients and SHAP will be used on each dataset to generate both clean and adversarial explanations. Algorithms \ref{alg:mult} and \ref{alg:add} (See Appendix) provide procedures to generate multiplicative and additive adversarial examples respectively. Table \ref{table:mlp_performance} provides the train and test performance of an MLP on each dataset.

The following describes how we generate an adversarial point for MNIST. Recall our assumption that the number of neurons $r$ is less than the dimension $n$. Let $\lbrace k^{1}, \ldots, k^{n-r} \rbrace$ be a basis for $\ker(\mathbf{W}_{1})$ and $x \in \R^{n}$ be some point. Choose $y \in \R^{n}$ such that $F(x) \neq F(y)$ and construct the set $I = \lbrace i \in [1,n] \mid x_{i} > 0 \rbrace$, here we are assuming the images are normalised between 0 and 1 and $|I| \leq n-r$. If $|I| > n-r$ one can repeat basis vectors. We construct the adversarial point via the following linear combination of a basis for $\ker(\mathbf{W}_{1})$:
\begin{equation}
\label{eqn:adv_point_mnist}
    \Tilde{x} = \sum_{i \in I} y_{i}k^{i}.
\end{equation}

Similarly for the Wisconsin breast cancer and network traffic dataset one can take the set $I$ to be some subset of features in the input. Using a basis of $\ker(\mathbf{W}_{1})$ one cannot construct a linear combination for an arbitrary point in $y \in \R^{n}$, unless $\Im(\mathbf{W}_{1})$ were trivial. The intuition for the above construction is that $\Tilde{x}$ will approximately resemble $y$ whilst being an element of the kernel.

\subsection{MNIST}

\begin{figure*}[!t]
  \centering
  \includegraphics[width = \linewidth]{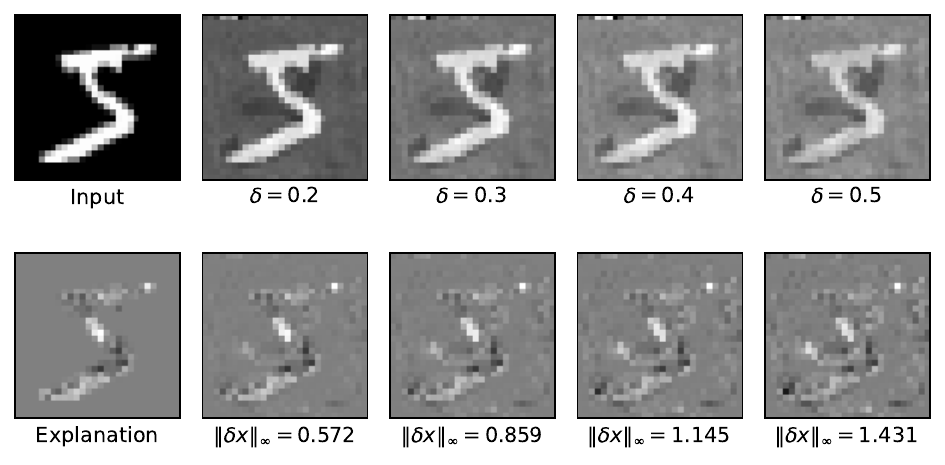}
  \caption{\textbf{Top:} Input of a digit 5 from MNIST. Adversarial point $\Tilde{y} = y + \delta x$, where $x$ is of the form in equation (\ref{eqn:adv_point_mnist}) to resemble a 3-digit and $y$ is a 5-digit. \textbf{Bottom:} Clean integrated gradients explanation followed by adversarial explanation with increasing $\delta$. The error between the clean point $y$ and adversarial point $\Tilde{y}$ is $\|\delta x \|_{\infty}$.}
  \label{fig:mnist_alg_attack}
\end{figure*}
Data in MNIST are 28x28 gray-scale images in 10-classes. Flattened into a vector for MLP input, images are $n = 784$ dimensions. We use a 2-layer MLP with $r = 16$ neurons per layer and ReLU activation. Note that $n > r$ meeting the condition of Corollary 1 \cite{alg_attacks_icmlc}. Furthermore we note that by the rank-nullity theorem $\dim(\ker(\mathbf{W})) = n - r$. We take $\varepsilon = 1$ to be the error threshold for MNIST. Pixel values in MNIST are normalised between 0 and 1, hence an error greater than 1 indicates the data is outside of the distribution of gray-scale images.

We see in Figure \ref{fig:mnist_alg_attack} the clean and adversarial input and the corresponding explanations. The adversarial input has the same classification as the clean input. As $\delta$ is increased, one achieves better adversarial explanations with the trade-off of the error threshold $\|\delta x \|_{\infty}$ increasing.

\subsection{Wisconsin Breast Cancer Dataset}
\begin{figure}[!t]
    \centering
    \begin{subfigure}[b]{0.5\textwidth}
        \centering
        \includegraphics[width=\linewidth]{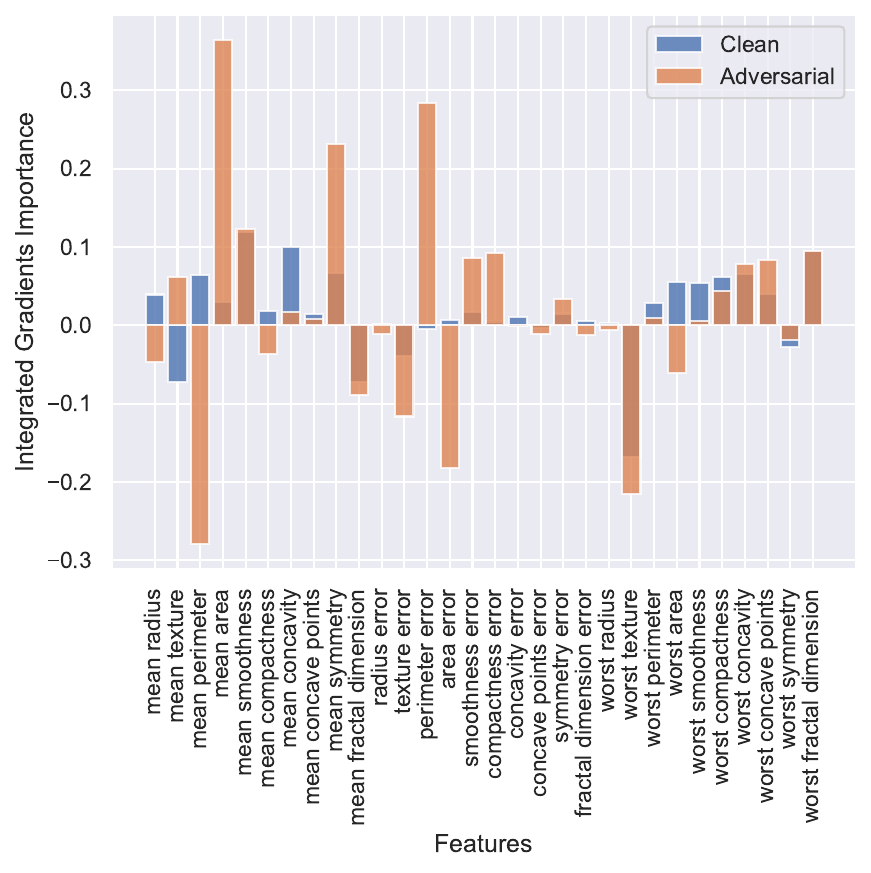}
    \end{subfigure}\hfill
    \begin{subfigure}[b]{0.5\textwidth}
        \centering
        \includegraphics[width=\linewidth]{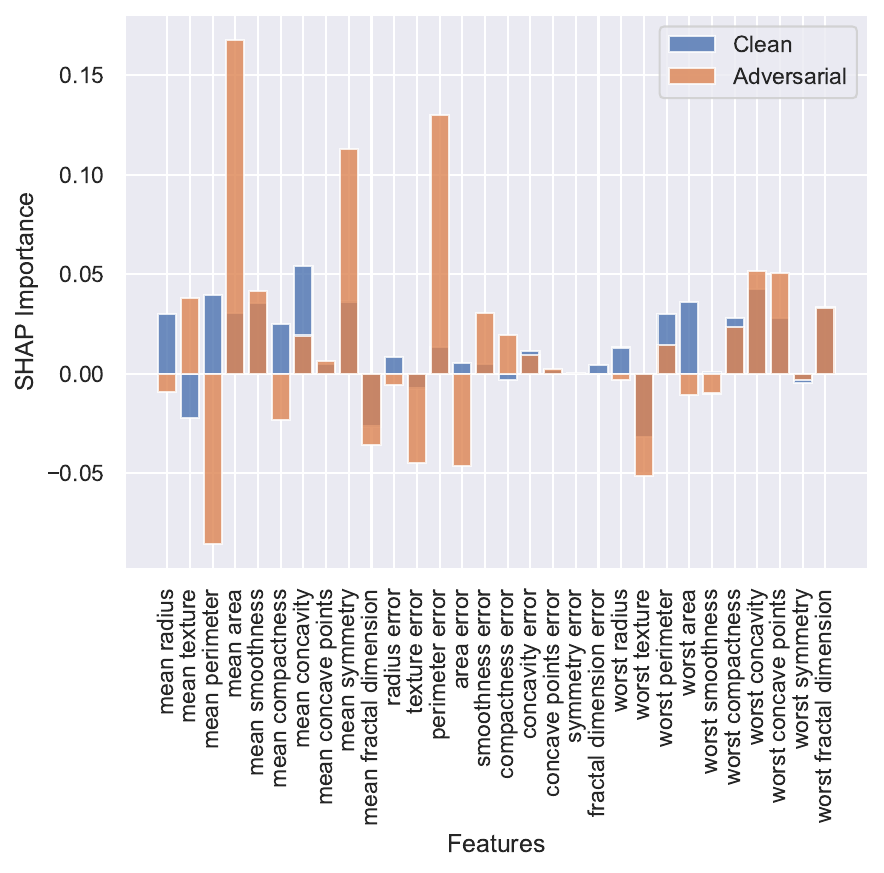}
    \end{subfigure}
    \caption{Clean and adversarial explanation of a `malignant' point in Wisconsin breast cancer dataset. \textbf{Left:} Integrated gradients feature importance. \textbf{Right:} SHAP feature importance.}
    \label{fig:breast_cancer_alg_attack}
\end{figure}

\begin{figure}[!t]
    \centering
    \begin{subfigure}[b]{0.5\textwidth}
        \centering
        \includegraphics[width=\linewidth]{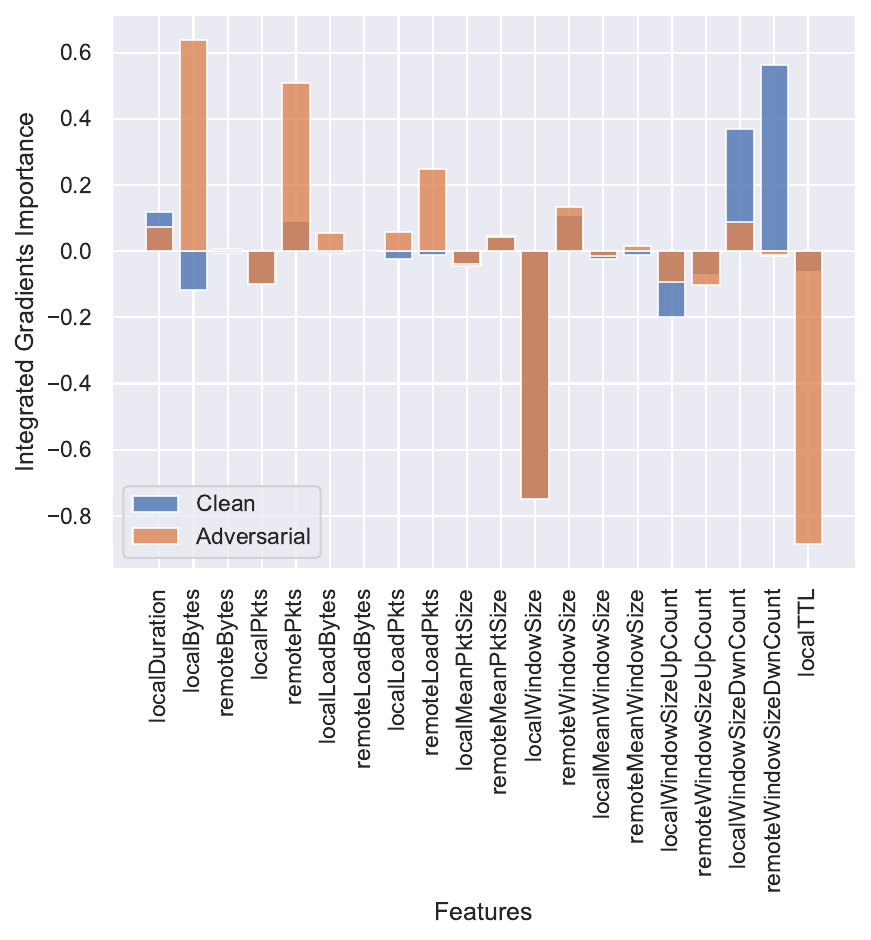}
    \end{subfigure}\hfill
    \begin{subfigure}[b]{0.5\textwidth}
        \centering
        \includegraphics[width=\linewidth]{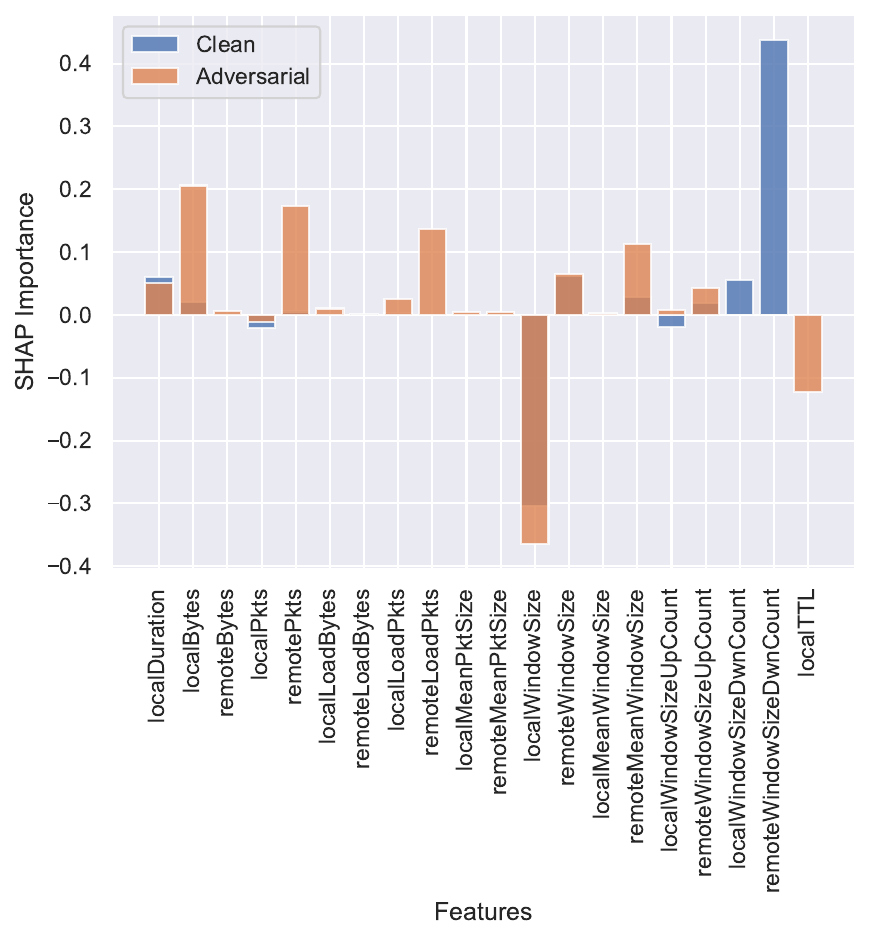}
    \end{subfigure}
    \caption{Clean and adversarial explanation of a point in the network traffic dataset classified as `Android'. \textbf{Left:} Integrated gradients feature importance. \textbf{Right:} SHAP feature importance. The error is $\|\delta x\|_{\infty} = 1$.}
    \label{fig:testbed_alg_attack}
\end{figure}

The Wisconsin breast cancer dataset is a standard binary classification dataset. The dataset consists of $n = 30$ statistical features of fine needle aspirates of breast masses with labels of either malignant or benign. We train a 2-layer MLP for classification with $r = 16$ neurons. Table \ref{table:mlp_performance} provides the train and test performance. 

Figure \ref{fig:breast_cancer_alg_attack} provides integrated gradients and SHAP feature importances of a clean and adversarial point classified as `malignant'. We see that the adversarial explanation places importance orthogonal of the clean point such as mean perimeter, mean radius, mean texture and worst area. The adversarial explanation in Figure \ref{fig:breast_cancer_alg_attack} places high importance on low important features in the clean point. Neural networks are increasing used in the medical domain to guide diagnosis. Explainability models help medical practitioners understand the reasoning behind the model. The adversarial attacks on the Wisconsin breast cancer dataset highlight the significant impact an adversary can have in the medical domain.

\subsection{Network Traffic of Mobile Devices}

The network traffic dataset \cite{simpson2023testbed} consists of two 40-hour captures of labelled network traffic from five mobile devices (3 Android and 2 iOS). The dataset has 36 flow features (20 numerical, 16 categorical). The network traffic is labelled by both the mobile application in use (e.g. Facebook Messenger, YouTube and Snapchat) and operating system (Android or iOS). Both application and operating system classification are critical in cyber-security to ensure quality of service and situational awareness. We train a 2-layer MLP with $r = 8$ neurons to perform operating system classification on $n = 20$ numerical flow features. Table \ref{table:mlp_performance} provides train and test performance. In Figure \ref{fig:testbed_alg_attack} we provide an algebraic adversarial attack on explanations provided by integrated gradients and SHAP on an `Android' traffic flow. We see in Figure \ref{fig:testbed_alg_attack} that the adversarial attack provides explanations orthogonal to the clean point. Adversarial explanations can obfuscate the network analysts understanding of the network traffic.

\section{Conclusions and Future Work}
In this work, we introduced the notion of algebraic adversarial examples for explainability models. We demonstrated that geometric deep learning provides the framework to study a certain symmetry group which provides an algebraic way of constructing adversarial examples. The symmetry group for adversarial examples was computed for three common deep learning architectures: MLP, CNN and GCN. We provided the conditions under which one can generate adversarial examples for equivariant path methods, neural conductance and additive feature methods. We further demonstrated that the extent one can achieve adversarial examples is bounded above by a scalar factor of the error threshold $\varepsilon$. Algebraic adversarial examples were experimentally tested on two well-known and one real world dataset, demonstrating the validity of our theoretical results. In future work, we seek to investigate other symmetry groups which lead to adversarial explanations. Furthermore, we will investigate the appropriate error threshold $\varepsilon$ on each dataset.

\section*{Acknowledgements}
The Commonwealth of Australia (represented by the Defence Science and Technology Group) supports this research through a Defence Science Partnerships agreement. Lachlan Simpson is supported by a University of Adelaide scholarship.

\bibliographystyle{IEEEtran}

\bibliography{arvix_draft}

\newpage
\appendix
\section{Appendix}
\label{appendix:lie-alg}

\subsection*{Pseudo-code for Algebraic Adversarial Attacks}

  \begin{algorithm}
  \caption{Compute Multiplicative Adversarial Example}\label{mult_alg}
  \begin{algorithmic}[1]
  \Require Point: $x \in \R^n$, weight matrix: $\mathbf{W}_{1}$
      \For{$y \in \ker(\mathbf{W}_{1})$}
        \State $\mathbf{B} \gets \mathbf{B} + x^ty-y^{t}x$
      \EndFor
      \State $t \gets \frac{1}{\|\mathbf{B}\|_{\infty}}\log(\frac{\varepsilon}{\|x\|_{\infty}}+1)$\Comment{Ensure $\|\Tilde{x} - x\|_{\infty} \leq \varepsilon$.}
       \State $\mathbf{A}_{(xy)} \gets \exp(t\mathbf{B})$\Comment{Maps from Lie algebra $\mathfrak{g}$ to symmetry group $G$.}

      \State \Return $\Tilde{x} = \mathbf{A}_{(xy)} \cdot x$\Comment{Multiplicative adversarial example.}
  \end{algorithmic}
    \label{alg:mult}
\end{algorithm}
  \begin{algorithm}
  \caption{Compute Additive Adversarial Example}\label{add_alg}
  \begin{algorithmic}[1]
  \Require Point: $x \in \R^{n}$, weight matrix $\mathbf{W}_{1}$
    \State Choose $y \in \ker(W)$
    \State $\delta \gets \frac{\varepsilon}{\|y\|_{\infty}}$\Comment{Ensure $\|\Tilde{x} - x\|_{\infty} \leq \varepsilon$.}
      \State \Return $\Tilde{x} = x+  \delta y$\Comment{Additive adversarial example.}
  \end{algorithmic}
      \label{alg:add}
\end{algorithm}

\subsection*{Lie Groups and Lie Algebras}

 We first provide basic definitions and examples related to matrix Lie groups and their actions on vector spaces. We refer the reader to \cite{ab_algebra_gallian} and \cite{Hall} respectively, for introductions on Lie theory and abstract algebra.

Let $G$ be a set, and let $* : G \times G \to G$, $(a, b) \to a * b$. We recall that the pair $(G, *)$, or simply $G$, is a group if it satisfies the following conditions:
\begin{itemize}
    \item[1.] There exists an element $e \in G$ such that $e * a = a * e = a$ for all $a \in G$.
    \item[2.] For each element $a \in G$, there an element $a^{-1} \in G$ such that $a * a^{-1} = a^{-1} * a = e$. 
\end{itemize}

In the literature, the map $*$ is usually called group operation, or group multiplication. The element $e$ is referred to as the identity of the group, and $a^{-1}$ the inverse of $a$. In the remainder of this work, we will omit the group operation an write $ab$ or $a + b$, depending on the context.

The most important examples of groups for this work are the \textit{complex general linear group}
\begin{equation*}
        \GL_{n}(\C) : = \lbrace g \in \gl_{n}(\C) \; : \; \det(g) \neq 0 \rbrace,
\end{equation*}
and the \textit{real general linear group}
\begin{equation*}
        \GL_{n}(\R) : = \lbrace g \in \gl_{n}(\R) \; : \; \det(g) \neq 0 \rbrace.
\end{equation*}
As sets, they are complex and real $n \times n$ matrices, respectively and, in both cases, they are groups with the usual matrix multiplication as group operation. The group identity corresponds to the identity matrix and group inverses are simply inverse matrices.
\begin{definition}
    Let $G$ be a subgroup of $\GL_{n}(\C)$. We will say that $G$ is a matrix Lie group if it is a closed subset of $\GL_{n}(\C)$. 
\end{definition}

Clearly, $\GL_{n}(\R)$ is a closed subgroup of $\GL_{n}(\C)$, since all convergent sequences of elements of $\GL_{n}(\R)$ converge to an element in $\GL_{n}(\R)$. In what follows, we will only consider matrix Lie groups contained in $\GL_{n}(\R)$. Below we provide two examples of matrix Lie groups which will be useful in the following sections.

Firstly, let us consider $\R^{n}$ with its usual addition as a group operation. It is indeed a group since, for any two pair of vectors $u, v \in \R^{n}$, we have $u * v = u + v \in \R^{n}$, $0$ is the group identity and $u^{-1} = - u$ is the inverse of $u$. To verify it is a matrix Lie group, we note that any $u \in \R^{n}$ can be represented by a matrix in $\GL_{n+1}(\R)$, of the form
\[
\begin{pmatrix}
    \Id & u \\
    0 & 1
\end{pmatrix},
\quad \Id \in \GL_{n}(\R), \quad u \in \R^{n},
\]
where $u$ is considered as a column vector.

A more sophisticated example is the \textit{real affine group} $\Aff_{n}(\R)$. It is a combination of both, $\GL_{n}(\R)$ and $\R^{n}$, which is given by the semi-direct product $\Aff_{n}(\R) : = \GL_{n}(\R) \ltimes \R^{n}$, with group operation given by
\begin{equation*}
    (g, u) * (h, v) := (gh, gv + u), \quad g, h \in \GL_{n}(\R), \quad u, v \in \R^{n}.
\end{equation*}
It is not difficult to verify that $(\Id, 0)$ is the group identity and that $(g, u)^{-1} = (g^{-1}, - g^{-1}u)$. By representing any element $(g, u) \in \Aff_{n}(\R)$ as a matrix of the form
    \[
    \begin{pmatrix}
        g & u \\
        0 & 1
    \end{pmatrix},
    \quad g \in \GL_{n}(\R), \quad u \in \R^{n}. 
    \]
    where $u \in \R^{n}$ is viewed as a column vector. It is not difficult to verify that $\Aff_{n}(\R)$ is in fact a subgroup of $\GL_{n+1}(\R)$, and therefor a matrix Lie group.

\begin{definition}\label{def:group_action}
 Let $G$ be a group and $X$ be a set. A left group action of $G$ in $X$ is a mapping $G \times X \to X$, $(g, x) \mapsto g \cdot x$ satisfying the following conditions:
    \begin{itemize}
        \item[1.] $e \cdot x = x$ for all $x \in X$.
        \item[2.] $g \cdot (h \cdot x) = gh \cdot x$ for all $g, h \in G$ and $x \in X$. 
    \end{itemize}
\end{definition}

Before providing examples of group actions, we remark that an action of a group $G$ on a set $X$ naturally induces an action in the set $F(X, Y)$, consisting of functions from $X$ to a set $Y$. For any $F \in F(X, Y)$, the action of $G$ in $F \in F(X, Y)$ is defined by
\begin{equation}\label{eqn:action_functions}
(g \cdot F)(x) = F(g^{-1} \cdot x), \quad \text{for all} \; x \in X.
\end{equation}
To verify that equation (\ref{eqn:action_functions}) is indeed a group action, we note that the first condition in Definition \ref{def:group_action} is trivially satisfied since
\[
(e \cdot F) (x) = F (e^{-1} \cdot x) = F(e \cdot x) = F(x), \quad \text{for all} \; x \in X.
\]
For any two $g, h \in G$ we have that
\[
(g \cdot (h \cdot F)) (x) = (h \cdot F)(g^{-1} \cdot x) = F(h^{-1} \cdot (g^{-1} \cdot x)) = F(h^{-1}g^{-1} \cdot x) = F((gh)^{-1} \cdot x) = (gh \cdot F)(x),
\]
and thus the second condition holds.

In this work we study the group actions when $G = \R^{n}$, $\GL_{n}(\R)$ or $\Aff_{n}(\R)$, $X = \R^{n}$ and $F \in F(\R^{n}, \R^{m})$ is a neural network.

\begin{example}
			$G = \R^{n}$ acting by translations in $\R^{n}$ and $F(\R^{n}, \R^{m})$:
			\[
			u \cdot x = x + u, \quad \quad (u \cdot F)(x) = F(u^{-1} \cdot x) = F(x - u).
			\]
		\end{example}
		\begin{example}
			$G = \GL_{n}(\R)$ acting by linear transformations of $\R^{n}$:
			\[
			g \cdot x = gx, \quad \quad (g \cdot F)(x) = F(g^{-1} \cdot x) = F(g^{-1}x).
			\]
		\end{example}
	
    Combining both groups actions from above, we obtain the action of the real affine group. The affine group allows acting on the neural network by dilations, rotations and translations to the input; a standard pre-processing step in machine learning tasks.
		\begin{example}
			$G = \Aff_{n}(\R)$ acting by affine transformations of $\R^{n}$:
			\[
			(g, u) \cdot x = gx + u, \quad ((g, u) \cdot F)(x) = F((g, u)^{-1} \cdot x) = F(g^{-1}(x - u)).
			\]
		\end{example} 

The Euclidean group $\mathrm{E}_{n}$ is the semi-direct product of orthogonal matrices $\On_{n}$ (rotations) and translations given by $\R^{n}$.

\begin{equation*}
    \mathrm{E}_{n} \coloneqq \On_{n} \ltimes \R^{n}.
\end{equation*}

Oftentimes it is difficult in practice to work with Lie groups. Lie Algebras provide a way to study Lie groups by a vector space. First we provide the definition of a Lie algebra and then the correspondence between Lie groups and Lie algebras via the exponential map.

\begin{definition}
    Let $\lie{g}$ be a real vector space and let $[\cdot, \cdot] \colon \lie{g} \cross \lie{g} \to \lie{g}$ be an $\mathbb{\R}$-bilinear skew-symmetric form. We will say that the pair $(\lie{g}, [\cdot, \cdot])$, or simply $\lie{g}$, is a real Lie algebra if $[ \cdot, \cdot ]$ satisfies the Jacobi identity:
    \begin{equation*}
         [x,[y,z]]+[y,[z,x]] + [z,[x,y]] = 0 \text{ for all } x,y,z \in \lie{g}.
\end{equation*}

\end{definition}

    Matrix Lie groups and algebras relate to each other via the exponential map. Recall that for a given $A \in \gl_{n}(\R)$, the exponential of $A$ is defined as the power series
    \begin{equation*}
        \exp(A) = \sum^{\infty}_{n = 0} \frac{A^{n}}{n!}.
    \end{equation*}

\begin{definition}\label{def:group_algebra}
    Let $G$ be a Lie subgroup of $\GL_{n}(\R)$. We will say that the vector space
    \begin{equation*}
        \lie{g} : = \lbrace A \in \gl_{n}(\R) : \exp(tA) \in G \text{ for } t \in \R \rbrace,
    \end{equation*}
    is the Lie algebra of $G$, where the bracket is given by the commutator of matrices, i.e. $[A,B] = AB - BA$ for all $A,B \in \lie{g}$.
\end{definition}

We remark that image of $\exp$ on $\gl_{n}(\R)$ is $\GL^{+}_{n}(\R) : = \lbrace g \in \GL_{n}(\R) \; : \; \det(g) > 0 \rbrace$. For more details we refer to \cite[Section 3.3]{Hall}.

\begin{proposition}[Polar decomposition]
\label{prop:polar_decomp}
    Let $g \in \GL_{n}(\R)$. Then there exists $Q \in \On_{n}$ and $S \in \Sym_{n}(\R)$ such that
    \[
    g = Q \; \exp(S).
    \]
\end{proposition}

\begin{proof}
    See \cite{Hall} Proposition 2.19.
\end{proof}

\end{document}